\documentclass[10pt, a4paper]{article}

\usepackage[]{lrec-coling2024}

\usepackage{graphicx}
\usepackage{booktabs}
\usepackage{subcaption}

\title{The Constant in HATE: Analyzing Toxicity in Reddit across Topics and Languages }

\name{Wondimagegnhue Tsegaye Tufa, Ilia Markov,  Piek Vossen} 

\address{Vrije Universiteit Amsterdam \\
         De Boelelaan 1105, 1081 HV Amsterdam, The Netherlands \\
         \{w.t.tufa, i.markov, p.t.j.m.vossen\}@vu.nl\\}

\abstract{
Toxic language remains an ongoing challenge on social media platforms, presenting significant issues for users and communities. This paper provides a cross-topic and cross-lingual analysis of toxicity in Reddit conversations. We collect 1.5 million comment threads from 481 communities in six languages: English, German, Spanish, Turkish, Arabic, and Dutch, covering 80 topics such as Culture, Politics, and News. We thoroughly analyze how toxicity spikes within different communities in relation to specific topics. We observe consistent patterns of increased toxicity across languages for certain topics, while also noting significant variations within specific language communities.
 \\ \newline }
\begin{document}
\maketitleabstract
\section{Introduction}

Social media platforms have witnessed remarkable growth in their user base and significance as a means of communication. These platforms allow individuals to share whatever they wish, presenting diverse viewpoints that range from enlightening to objectionable and everything in between. As a side effect, platforms often provide a breeding ground for toxic content, such as instances of abuse and hate speech, resulting in adversities for online users \citep{fortuna2018survey}. Outside the confines of social media, this toxic content influences real-world dynamics. These are often manifested in instances of violence and crimes targeting minority groups \citep{mathew2021hatexplain}.
\begin{figure}[htbp]
  \centering
  \resizebox{0.5\textwidth}{!}{%
    \includegraphics{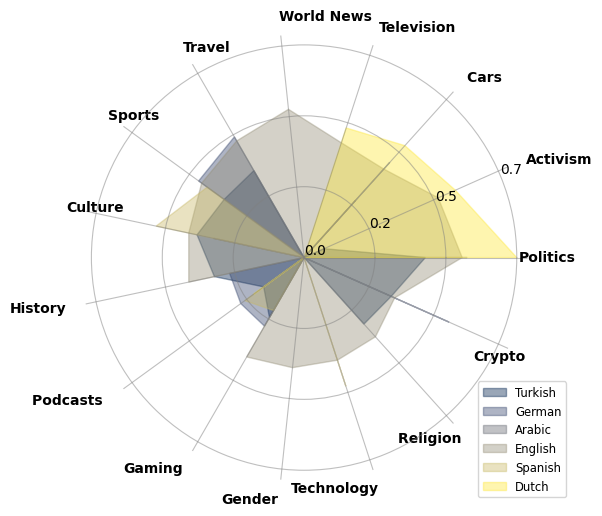}
  }
  \caption{Comparison of toxicity levels in Reddit discussions across different topics and languages. The scores represent the toxicity density, the proportion of toxic comments within each topic. Each line illustrates the toxicity density for a specific language within a particular topic.}
  \label{fig:main}
\end{figure}
The detection of toxic content has emerged as a progressively significant subject of investigation within the field of Natural Language Processing (NLP). Active research in this area focuses on creating datasets that cover different aspects of toxic content \citep{mathew2021hatexplain,vidgen2021introducing,sachdeva-etal-2022-measuring}, or methods that rely on these datasets to analyze toxic content or train toxic language classification systems \citep{van2018challenges,radfar2020characterizing,gevers2022linguistic,markov-etal-2022}. 

While many existing studies focus on classifying whether a given text is toxic and why, the context in which such inappropriate content arises is less explored \cite{zhou-etal-2023-cobra,sap-etal-2019-risk}. 

In Reddit, a specific discussion often turns toxic when the topic of discussion is sensitive to a particular user. Participants of such discussions with opposing views engage in unhealthy debates, which can quickly escalate.  A sensitive topic may evoke strong emotions, making participants use offensive remarks. This emotional intensity, combined with Reddit's anonymity, can lead to personal attacks and offensive language use. Additionally, the platform's upvote and downvote system can reinforce popular opinions, creating echo chambers dominated by extreme viewpoints. Consequently, the type of topic being discussed might be a central factor for its potential descent into toxicity.

Reddit, as a social platform, has gained significant attention in the area of toxic language research \citep{baumgartner2020pushshift}. The platform offers easy access to data collection in comparison to platforms such as Facebook and Twitter \citep{baumgartner2020pushshift}. It is also reported that there is a significant inclination towards the use of language considered toxic and offensive \citep{demszky2020goemotions}. This characteristic makes Reddit an ideal platform for studying how toxic language is manifested in various communities. Central to Reddit's structure are subreddits, proxies to communities comprising members who share mutual interests, such as political viewpoints or leisure pursuits. User interaction frequently occurs within these community boundaries around a particular topic. 

The prevalence of toxic language on platforms like Reddit has been widely researched. These studies focused on aspects such as individual user comments and posts~\citep{kumar2023understanding,hiaeshutter2022language}, community-level conversations \citep{farrell2019exploring}, or the behavior of the users \citep{urbaniak2022namespotting}. 


In this study, we view toxicity on Reddit from a broader contextual perspective encompassing topic, community, and language. We are specifically interested in how toxicity develops within communities in relation to topics. For this reason, we collect not only specific comments that are likely to be toxic but also the subthreads in which such comments occur, which may also exhibit more nuanced cases of stereotypical targeting, implicit hate speech, irony, and sarcasm within communities. We, therefore, collect conversation threads from Reddit spanning different languages, communities, and topics.\footnote{The data and our analysis are available following the US and EU FAIR use principles and according to the license conditions of Reddit on source data. The GitHub repository can be accessed here: \url{https://github.com/cltl/Reddit_topic/tree/main}}

Our analysis shows that toxicity in Reddit conversations strongly depends on the topic of discussion. As shown in Figure \ref{fig:main}, certain topics show high toxicity in most of the target languages (e.g., Politics, Sports).  In our monolingual analysis, we show that topics that would normally be considered neutral, such as History and Gaming, still have the potential to trigger toxicity.  We also observe measurable differences in the toxicity of certain topics across languages.
The result of our analysis can be used in several ways. Social media moderators can use the insights from our study for more effective content moderation. Since toxic content is more common in some topics than others, focusing on toxic-prone topics can be more efficient for filtering inappropriate content. It is also important to consider cultural differences. Our analysis shows that topics considered less toxic in one language are more prone to generate toxicity in another. In terms of training models for automatic content moderation, topic and language can be considered part of the context of a comment. This context information can be used in model training for more accurate detection of toxic content. 

In summary, our contributions are:
\begin{itemize}
    \item We collect 1.5 million comment threads from 481 communities in six languages.
     \item We explore the relationship between toxicity and topics of conversation in mono-lingual and cross-lingual settings across different Reddit communities.
    \item We compare and contrast three distinct approaches to measure toxicity.
\end{itemize}

\section{Related Work}
The social media landscape has become a dynamic arena where users and groups interact, share their diverse viewpoints, and communicate. Within this theme, the occurrence and consequences of toxic language have garnered substantial attention from researchers across various disciplines, such as social sciences, political science, and NLP.
Here, we use toxic language as an umbrella term similar to \citet{sharma2022detecting}, broadly comprising hate speech, offensive language, abusive language, propaganda, cyberbullying, and cyber-aggression. In this section, we provide an overview of studies that analyze one or more aspects of toxic language in social media settings from user and community perspectives. 

\paragraph{Comment and post analysis} \citet{kumar2023understanding} provide an extensive study of the behavior of accounts on Reddit that post toxic content. The study shows that although accounts engaging in abusive behavior make up less than 4\% of Reddit's total users, they are responsible for generating 33\% of all comments posted on the platform.
\citet{mall2020four} also explore similar user behavior analysis through a temporal analysis of user toxicity and show that the typical behavior of toxic users is switching between toxic and non-toxic commenting. Similar work by\citet{hiaeshutter2022language} studies the relationship between major political events and hostility in a discussion using language analysis. The findings indicate that U.S. political events led to heightened hostility and increased negativity in Reddit discussions. \citet{urbaniak2022namespotting} study correlation between username toxicity and toxic behavior of these users on Reddit. Users who have toxic usernames generate a greater amount of toxic content compared to those with neutral usernames.  

\paragraph{Community analysis} 
\citet{farrell2019exploring} constructed specific sets of lexicons to systematically study the changes in language use within Reddit communities known for misogynistic discussions.
In the context of discussing negative interactions, as highlighted by \citet{urbaniak2022namespotting} in their work on "namespotting", \citet{kumar2018community} present findings that align with this observation, showing that a small percentage of Reddit communities are responsible for the majority of negative interactions on the platform.
\citet{radfar2020characterizing} explore toxicity in Twitter from the user relation perspective and show that tweet exchanges between users without any connection are three times more prone to toxicity than interactions involving mutual friends. 

\paragraph{Toxic language resource} There are various lexical resources for different languages that define offensive words. Such resources include HurtLex \citetlanguageresource{bassignana2018hurtlex}, MOL  \citetlanguageresource{vargas-etal-2021-contextual}, DALC  \citetlanguageresource{caselli2021dalc}, and Hatebase  \cite{website:hatebase}. HurtLex is a lexicon that covers 50 languages and is divided into 17 categories, including ethnic slurs and derogatory terms, among others. MOL is a lexicon of abusive language annotated with contextual information. It covers English, Spanish, French, German, and Turkish. DALC is a Dutch lexicon of abusive words manually annotated from a Twitter corpus. Hatebase is a crowdsourced resource of hate speech lexicons. Though the Hatebase project was discontinued, the website can be accessed as a browsable archive. NRC lexicon  is a manually annotated emotion lexicon for English \citelanguageresource{Mohammad13}. It includes basic emotions and sentiments, as well as their associated emotions.  We specifically consider the NRC lexicon because our interest lies in understanding toxicity in a broader sense. This includes identifying negative sentiments, which are crucial for recognizing instances of implicit hate speech.

\paragraph{Measuring toxicity } For quantifying the toxicity of a comment, a widely used approach is Google's Perspective API \cite{lees2022new} and Detoxify \citep{Detoxify}. Perspective is trained on comments to capture the toxicity of a text in various contexts \citep{salminen2020topic}. It supports the detection of toxicity, insult, profanity, identity attacks, threats, and sexually explicit content. It covers multiple languages, including Arabic, English, German, Dutch, and Spanish. Detoxify is trained on the jigsaw challenges dataset for toxic comment classification \citep{Detoxify}. It supports English, French, Spanish, Italian, Portuguese, Turkish, and Russian. 

\paragraph{Topic and language analysis} 
A study by \citet{salminen2020topic} explores the relationship between toxicity and news topics. The results show that discussions related to racism, Israel-Palestine, and war exhibit higher toxicity in comments. It also shows instances of a typically less toxic topic that becomes more toxic when politics and religion are involved. A similar analysis by \citet{hilte2023haters} analyzes profiles of users who post toxic content in different languages such as English, Dutch, Slovenian, and Croatian. Both of these works are similar to our work in using topics to analyze toxicity. In comparison, our work can be considered complementary as we include a broader range of topics and more languages in our analysis.

\section{Methodology}
\subsection{Data source}
We collected a total of 1.5 million comments in 80 topics and six languages. Each of the comments includes a timestamp, an anonymized username, the subreddit, the topics of the subreddit, the submission in which the comment was posted, the submission title, and the body. We also include graph data that enables us to reconstruct the thread structure. Ultimately, we are interested in analysing subthreads that have a high chance of exhibiting both implict and explicit toxic behaviour.

We anonymized the author's personal information according to GDPR regulations. We first identify user names from the author name attribute of our collected metadata. We then replaced each identified user name with a unique and non-descriptive identifier consisting of a random string and numerical code to remove any connection to the individual.

\begin{table}[htbp]
\centering 
    \begin{tabular}{|c|c|}
         \toprule
        \#Language & 6 \\
        \# Topics & 80 \\
        \# Communities & 481 \\
        \# Submissions & 39,249 \\
        \#Unique Users & 511,464 \\
         \# Comments & 1,543,272 \\
        \bottomrule
    \end{tabular}
    \caption{Statistics for the collected data. Communities refer to the subreddit. Threads are all the comments under the same submissions or posts.}
    \label{tab:main_stat}
\end{table}


\subsection{Data collection and preprocessing}
We use PRAW\footnote{ https://praw.readthedocs.io/en/stable/index.html}, the Reddit Python package, to collect the data. We first extract lists of subreddits from the Reddit community ranking page.\footnote{ https://www.reddit.com/best/communities/1/} The website contains subreddits ranked by the number of subscribers. 

\paragraph{Language detection}
The Reddit API doesn't provide language information about the subreddit. To identify the language of a subreddit, we use Google Translate API to automatically classify the description of the subreddit into target languages. Since Reddit is predominately used in the English speaking communities, the most popular subreddits are in English. To create a balanced list of subreddits for each of our target languages, we create a new list from the initial list by sampling an equal number of subreddits per language. We then collect posts and comments from each subreddit. We query 100 popular submissions for each subreddit based on the upvote count. We then collected all the comments under these popular submissions. This initial list contains 178K subreddits. Table \ref{tab:main_stat} shows the main statistics of the collected data.

\paragraph{Topic identification}
In order to determine the topic of a specific subreddit, we employ a different approach. Since the Reddit API does not provide information about a subreddit's topic, we undertake a separate web crawl from the Reddit community ranking page. This allows us to associate each subreddit with its corresponding topic category.

\paragraph{Pre-processing}
We excluded comments that are either shorter than 15 characters or longer than 300 characters in length or comments which contains only emojis or punctuation. This decision aligns with previous research addressing the limitations of applying existing toxicity models to short, excessively long or noisy texts \citep{kumar2023understanding}. 

\begin{figure}[htbp]
  \centering
  \includegraphics[width=0.5\textwidth]{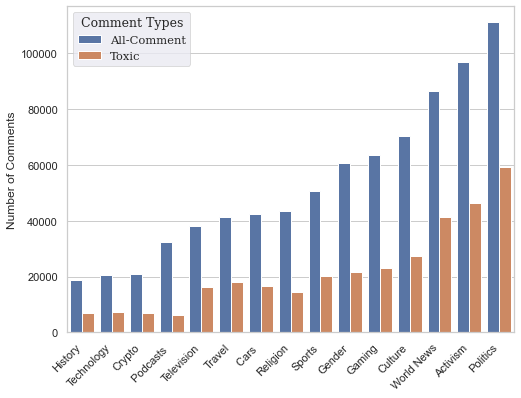}
  \caption{Distribution of toxic comments across topics based on the lexical-based approach. For visibility, we show the top 15 topics. Here, a comment is considered toxic if it contains at least one toxic word.}
  \label{fig:dist_comment_topic}
\end{figure}

\subsection{Toxicity scores}
We stress that our ultimate goal is to create a dataset across communities in which we can find subreddit threads with a high probability of exposing toxic language and, specifically, hate speech. This should contain cases of not only explicit but also implicit hate speech. Because it is more difficult to find implicit hate speech, we are interested in a method that has high recall of finding toxicity comments so that we can further analyse the subthreads in which these occur. To decide on a high-recall method, we conduct a manual assessment of three methods to identify comment toxicity, focusing on those with the broadest applicability to our target languages. These methods include the Perspective API, a lexicon-based approach, and OpenAI's GPT-4.

\subsubsection{Lexicon-based approach}
For the lexicon-based approach, we combine HurtLex, MOL, DALC, Hatebase, and NRC and build a binary classifier to score the toxicity of a comment.
For the NRC lexicon, we ignore the emotion layer and only used words that are associated with negative sentiment. If at least one toxic word is present in a comment, we consider it toxic. Lexicon-based approaches have shown to be robust when detecting toxic words in cross-domain settings \citep{schouten2023cross} and can easily be extended to other languages or adapted in the future. Our merged lexicon has 4,316 English, 7,041 Dutch, 1,831 Arabic, 2,782 Turkish, 2,903 Spanish, and 2,851 German words.

\subsubsection{GPT-4}
For GPT-4, we employ a simple zero-shot prompt to assign toxicity labels to a comment. We include a definition of what a toxic comment is in the prompt. We prompt GPT-4 to classify comments as toxic if it is hate speech, offensive language, abusive language, propaganda, cyberbullying, or cyber-aggression or non-toxic otherwise. Our prompt is "Review each comment and label it as toxic or non-toxic. To determine whether the comment is toxic if the comment falls into any of the following categories: hate speech, offensive language, abusive language, propaganda, cyberbullying, or cyber-aggression. If the comment aligns with any of these categories, label it as 'Toxic' in the label column. If the comment does not fit any of these categories, label it as non-toxic ". 

\subsubsection{Perspective API}
Perspective API is an out-of-the-box toxicity classifier from Google. The  API takes a comment as input and produces a score between 0 and 1 for different toxicity categories, such as threats, profanity, and identity attacks. Since we are interested in an aggregate score, we use the toxicity attribute to get a single score. Based on a recommendation from the API documentation, we use a threshold value of 0.75, and we consider a comment toxic if its toxicity score is higher than this threshold value.

\subsubsection{Expert annotation}
We conduct an expert annotation to identify the most effective method for detecting toxic comments. Our goal is to evaluate the performance of the identified approaches, particularly focusing on high recall. We randomly sampled 500 comments from each language from our dataset. We prepared  annotation guidelines with the definition of what kind of comment should be labeled as toxic. Our definition of toxic comment comprises hate speech, offensive language, abusive language, propaganda, cyberbullying, and cyber-aggression. We selected native speakers as subject matter experts. The annotators classified comments as toxic or not toxic based on the provided guidelines. We resolved questions and discrepancies through discussion. The languages covered in this paper include German, Turkish, Spanish, Dutch, Arabic, and English. 

\subsubsection{Thread toxicity}
We use this metric to compute the toxicity of a thread (instead of  single comment), where thread refers to all the comments that are part of a single submission. This analysis gives a more robust estimation of toxicity since a thread can have multiple comments from different users. To do this, we first reconstructed the thread structure of the comment from our dataset. We then filter threads with at least ten comments before computing the {thread toxicity.

\subsubsection{Topic Toxicity }
We define topic toxicity as the proportion of toxic comments on a specific topic relative to the total number of comments on that topic. We computed topic toxicity for each topic in the target language. 

\begin{table*}[ht]
\small
\centering
\label{tab:performance_metrics}
\begin{tabular}{@{}llccccccccccc@{}}
\textbf{} & \textbf{} &
\multicolumn{3}{c}{\textbf{Lexical}} &
\multicolumn{3}{c}{\textbf{Perspective}} &
\multicolumn{3}{c}{\textbf{Gpt-4}} \\
\cmidrule(lr){3-5} \cmidrule(lr){6-8} \cmidrule(lr){9-11}
& & \textbf{P} & \textbf{R} & \textbf{F1} &
\textbf{P} & \textbf{R} & \textbf{F1} &
\textbf{P} & \textbf{R} & \textbf{F1} & \textbf{Support}\\
\midrule
& Non toxic  & \underline{.90} & .62 & .74 & .88 & \underline{.98} & \underline{.93} & .87 & .87 & .87 & 1315\\ 
& Toxic      & .17 & \underline{.53} & .25 & .35 & .08 & .13 & .08 & .08 & .08 & 190\\
& Macro avg  & .53 & .57 & .49 & .61 & .53 & .53 & .48 & .48 & .48 & 1505\\
\midrule
DE        & Non toxic  & \underline{.97} & .46 & .63 & .96 & .93 & .95 & .94 & .94 & \underline{.94} & 240 \\
& Toxic      & .07 & \underline{.69} & .12 & .31 & .24 & .24 & .31 & .31 & .31 & 13\\
& Macro avg  & .52 & .58 & .37 & .58 & .62 & .59 & .47 & .47 & .47 & 253\\
ES        & Non toxic  & .88 & .46 & .61 & .82 & \underline{.99} & \underline{.88} & .86 & .86 & .86 & 178 \\
& Toxic      & .30 & \underline{.79} & .44 & .82 & .17 & .28 & .29 & .29 & .29 & 53 \\
& Macro avg  & .59 & .63 & .52 & .81 & .58 & .58 & .53 & .53 & .53 & 231 \\
NL        & Non toxic  & \underline{.97} & .38 & .55 & .94 & \underline{1.00} & \underline{.97} & .96 & .96 & .96  & 252\\
& Toxic      & .08 & \underline{.81} & .14 & .00 & .00 & .00 & .12 & .12 & .12  & 16\\
& Macro avg  & .52 & .60 & .35 & .47 & .50 & .48 & .54 & .54 & .54  & 268\\
AR        & Non toxic  & .88 & \underline{.94} & \underline{.91} & .86 & 1.00 & .92 & .86 & .86 & .86  & 457\\
& Toxic      & .41 & .24 & .30 & .00 & .00 & .00 & .00 & .00 & .00  & 75\\
& Macro avg  & .65 & .59 & .61 & .43 & .50 & .46 & .43 & .43 & .43  & 532\\
EN        & Non toxic  &\underline{.86} & .51 & .64 & .85 & \underline{.95} & \underline{.90} & .84 & .84 & .84  & 188\\
& Toxic      & .16 & \underline{.55} & .25 & .17 & .06 & .09 & .06 & .06 & .06  & 33\\
& Macro avg  & .51 & .53 & .45 & .51 & .50 & .49 & .47 & .47 & .47  & 221\\
TR        & Non toxic & .69 & .57 & .62 & - & - & - & .6 & \underline{.87} & \underline{.71}  & 180\\
& Toxic      & .49 & \underline{.61} & \underline{.54} & - & - & - & .37 & .12 & .18  & 120\\
& Macro avg  & .59 & .59 & .58 & - & - & - & .48 & .49 & .44  & 300\\
\bottomrule
\end{tabular}
\caption{Evaluation of Lexical-based approach and Perspective API. The first three rows show the aggregate result for all languages, followed by a language-specific breakdown. Here, we put '-' since Perspective doesn't support Turkish. We also exclude Turkish in the aggregate result of the first three rows.}
\label{tab:aggregate_evaluation}
\end{table*}

\section{Results and Analysis} 
In this section, we present the main findings. We divide our analysis into three parts. First, we compare the performance of the different methods in detecting toxic comments based on the test data we created. We then explore the relationship between toxicity and topics in aggregate and for each language separately.
Finally, we analyze how consistent a topic toxicity is across languages by comparing the toxicity results across the six languages covered in this paper.   

\subsection{Evaluation of approaches}
\label{sec:lexical_vs_perspective}
We present the result of the evaluation of the three approaches in Table~\ref{tab:aggregate_evaluation}. In the aggregate results, we observe a significant difference across the approaches. The lexical-based approach significantly outperforms both Perspective-API and GPT-4 in terms of recall of toxic comments (respectively .53, .08 and .08), whereas Perspective outperforms to the others in precision (.35 versus .17 lexical and .08 GPT-4). Similarly, the cross-lingual analysis shows that the lexical approach consistently has the highest recall in the toxic category, indicating that this approach is the most effective in identifying toxic content with high recall across languages. We do see some differences between languages, as the precision scores for Dutch and German using the lexical approach are significantly lower.

As we stated before, we prioritize recall over precision for our analysis because we want to maximize the probability that we find threads that exhibit explicit or implicit toxicity. Toxic comments are rare compared to non-toxic ones \cite{vidgen2020directions}. We aim to flag potential toxicity broadly on this first pass to ensure that any potential toxic content is not missed, accepting the false positives.

\subsection{Topic toxicity}\label{sec:topical_toxicity}
In this section, we analyze topic toxicity in aggregate. We first identify the top 15 topics from the 80 topics based on the number of comments. Figure \ref{fig:dist_comment_topic} shows the distribution of toxic and non-toxic comments for the top 15 topics. In the distribution, politics-related topics such as Politics, Activism and news-related topics like World News have a higher number of toxic comments. 
For a more accurate comparison of the toxicity of topics, we computed the topic toxicity for each topic as described in the methodology section. Since the topic toxicity is a normalized value, it is possible to directly compare this value across topics. 

Similar to the distribution, we found topics related to Politics and World News to have the highest topic toxicity. This is partially consistent with the results reported by \citep{salminen2020topic}, which shows that topics related to Politics and News are highly likely to generate toxic conversation. In contrast, we also observe high toxicity in less expected topics such as Travel and History.

\begin{figure}[htbp]
  \centering
  \resizebox{0.48\textwidth}{!}{%
    \includegraphics{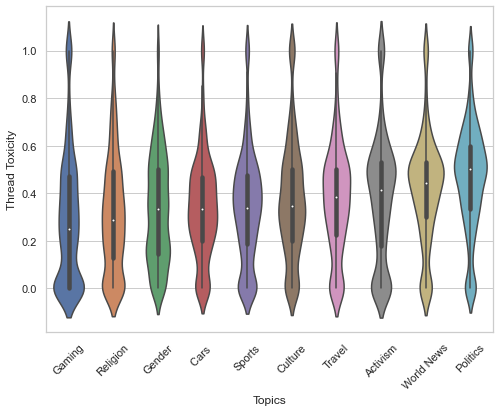}
  }
  \caption{Distribution of thread toxicity across topics. For visibility, we only show the top 10 topics. Plots are sorted by the mean value.}
  \label{fig:dist_thread_topic}
\end{figure}

\subsection{Distribution of toxic threads}
As described in the methodology section, we use thread toxicity for a more accurate estimation of toxicity. The thread toxicity provides an aggregate score rather than relying on the toxicity score of a single comment. In this analysis, we first group comments into different comment threads using the parent-child relationship of comments and submissions. We then compute the thread toxicity for each of the threads. Comment threads with more toxic comments will have a score close to 1, and threads with less toxic comments will have a value close to 0,
as shown in Figure \ref{fig:dist_thread_topic}. 

The y-axis represents the toxicity level, ranging from 0 to 1, and the x-axis shows different Reddit topics. Each violin shape provides a density estimate of the data at different toxicity levels. The wider a section of the violin, the higher the density of threads at that toxicity level. Here, we notice that Politics, World News, and Activism have a higher mean toxicity score and a greater number of threads with high toxicity scores. A dense concentration of toxic thread for Activism shows a broad dispersion, with a high density in the upper quartile, indicative of the potential contentiousness of discussions on this topic. In World News, while there is a significant central tendency around the median, a non-negligible spread towards the upper toxicity range is evident. Lastly, Politics is characterized by its extensive variance and significant density at the toxicity scale's lower and upper bounds. 

\begin{figure}[htbp]
  \centering
  \resizebox{0.5\textwidth}{!}{%
    \includegraphics{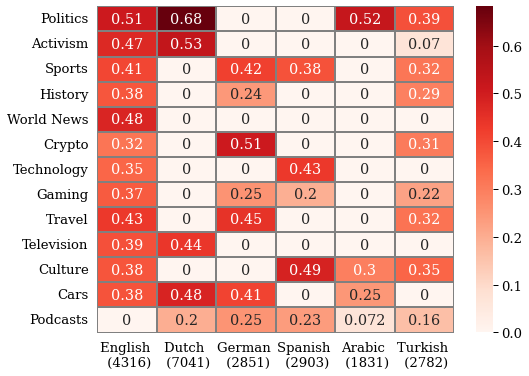}
  }
  \caption{Toxicity scores using the lexicon-based approach. The number under each language shows the total number of lexicon entries in that language.}
  \label{fig:all_heatmap}
\end{figure}

\begin{table*}[htbp]
\small
\centering
\begin{tabular}{l|l|l|l|l|l}
\toprule
Arabic & Turkish & Spanish & German & Dutch & English \\ 
\midrule
\textbf{Politics} & \textbf{Politics} & \textbf{Culture} & Crypto & \textbf{Politics} & \textbf{Politics} \\ 
\textbf{Culture} & \textbf{Culture} & Technology & Travel & \textbf{Activism} &  News \\ 
Cars  & Travel & \textbf{Sports} & \textbf{Sports} & Cars  & \textbf{Activism} \\ 
Podcasts  & \textbf{Sports} & Podcasts  & Cars & Television & Travel \\ 
\textbf{Activism} & Crypto & Gaming & Gaming & Podcasts  & \textbf{Sports} \\ 
\bottomrule
\end{tabular}
\caption{List of top five topics that have the highest topic toxicity score in each language. Topics that are toxic in more than two languages are shown in bold.}
    \label{tab:topic_toxicity_per_language}
\end{table*}

\subsection{Monolingual topic-toxicity}\label{sec:mono-linqual}
We compute the topic toxicity per language to analyze which topics stand out as more toxic than others in each language. Table \ref{tab:topic_toxicity_per_language} shows each language's top five toxic topics based on topic toxicity. Since the topic toxicity is a normalized value, we can use it to compare topic toxicity within and across languages. 

English comments have the highest toxicity in Politics and Activism. In terms of intensity, conversations related to politics and news have the highest toxicity. Similar to the aggregate result, this result is partially in line with  \citep{salminen2020topic}. Contrary to \citep{salminen2020topic}, discussions related to religion do not have high toxicity in our analysis. For Arabic conversations, partially similar to English, we observe high toxicity in discussions related to politics, such as Politics and Activism. We also observe high toxicity in discussions that involve Culture. In German, contrary to English and Arabic conversations, we observe high toxicity in more unexpected topics such as Crypto, Travel, and Cars. Similar to English and Arabic, Dutch conversation has the highest toxicity in political conversations. In Spanish, similar to German, the toxicity is concentrated in less expected topics such as Technology and Gaming.

In summary, conversations in Politics and Sport consistently show high toxicity in four out of the six target languages. We also observe high topical-toxicity patterns in Culture (Spanish, Arabic, and Turkish) and Gaming (Spanish and German). In the next section, we expand on a cross-lingual toxicity analysis for topics shared across the target languages.

\subsection{Cross-lingual toxicity analysis}
\label{sec:Cross-Linqual}

For cross-lingual analysis, we select topics that are shared by at least two languages. 
Figure~\ref{fig:all_heatmap} shows a Heatmap of toxicity across the selected topics and languages.\\
\subsubsection{Consistent toxicity in politics}
Politics, one of the cross-lingual topics shared by English, Dutch, Arabic, and Turkish, shows the most consistent toxicity in English, Dutch, and Arabic. In terms of intensity, we observe that it is more toxic in Dutch than in the other languages we analyzed. In general, we observe a similar pattern of toxicity with variation in intensity. 

\subsubsection{Diversity in toxic topics} While some languages like Dutch and Arabic show high toxicity in topics such as Politics and Activism, others like German demonstrate high toxicity in seemingly neutral topics like Crypto and Travel. The Spanish conversations tend to express stronger reactions when discussing culture and ethnicity. English and Turkish languages show a more diverse picture; comments in these languages display varied toxicity levels across multiple topics. This suggests that users in these languages have a broader range of subjects that elicit strong, potentially toxic responses. The results underscore the cultural and linguistic nuances in how different topics are perceived and discussed across languages.


\section{Conclusion}
Our findings support prior research emphasizing the relationship between topics and the toxicity of a comment. We broaden this correlation to encompass a broad range of topics and languages. In the aggregate analysis, we found conversations that involve politics and news to have the highest toxicity, which is partially consistent with the results reported by \citet{salminen2020topic}. In contrast, we also observe high toxicity in less-expected topics such as travel and history. In monolingual analysis, we demonstrate that conversations in Politics and Sports consistently show high toxicity in the majority of our target languages. We also observe such topical-toxicity patterns in Culture, Ethnicity, and Gaming. Furthermore, we observe major differences across languages in relation to the topics. Whether these differences also correspond to variations in community dynamics cannot be determined from the current data. Further investigation is required to answer to what extent these language communities actually discuss the same things within the broader topic clusters. In future research, we want to analyze the topics of the subreddits in more detail using entity recognition and topic classification in comparison to similar time frames to further compare the content across languages. Similar entities and topics in similar periods could be used as an indication of parallel discussion across communities that potentially exhibit different toxicity. Furthermore, we want to analyze the build-up of toxicity within the thread and also focus on the targets of such language and implicit hate speech instances in our dataset.

\subsection{Limitations}
We identify some limitations in our work. First, using topics to categorize a subreddit can oversimplify the rich nuances of a conversation that may take place in a particular community. Many conversations may not clearly fit into one topic, often overlapping with multiple topics. These conversations are also dynamic in nature, with threads evolving and branching into subtopics. A static categorization might not capture the fluidity of these discussions. The level of detail within a topic is another factor to think about, as certain topics can be overly general while others are highly specific. Finding the right balance between granularity and generality in categorization is challenging. The lexicons we use for computing the toxicity also have a limitation. The variation in the quality and quantity of lexicon items for each language might lead to results that favor certain languages over others.


\subsection{Ethical consideration}
In this paper, we use information collected from the Reddit platform, a public online platform where users post content and take part in discussions. We recognize and emphasize the importance of ethical considerations when handling and analyzing such datasets. Firstly, all data used were publicly accessible and did not involve any private or confidential information. We take all the necessary steps according to GDPR regulations to anonymize any identifiable user information to ensure privacy. Furthermore, we use the collected data strictly for research purposes, and no attempt was made to exploit, manipulate, or otherwise use the data in a manner that could harm or prejudice any individual or group. Any insights drawn from this work are based only on patterns in the data and should not be used to stereotype or make generalizations about specific groups or individuals. 

\subsection{Acknowledgements}
The research was supported by Huawei Finland through the DreamsLab project. All content represented the opinions of the authors,
which were not necessarily shared or endorsed by their respective employers and/or sponsors. We would like to thank our colleagues Thami Zabda, Lisa Beinborn, Selene Báez Santamaría, and Mekselina Do{\u{g}}anç for assisting us in the annotation task.
\nocite{*}
\section{Bibliographical References}\label{sec:reference}
\bibliographystyle{lrec-coling2024-natbib}
\bibliography{lrec-coling2024-example}
\section{Language Resource References}
\label{lr:ref}
\bibliographystylelanguageresource{lrec-coling2024-natbib}
\bibliographylanguageresource{languageresource}
\end{document}